\documentclass{article} % For LaTeX2e
\usepackage{iclr2025_conference,times}

\usepackage{amsmath,amsfonts,bm}

\def\eqref#1{equation~\ref{#1}}
\def\1{\bm{1}}

\DeclareMathAlphabet{\mathsfit}{\encodingdefault}{\sfdefault}{m}{sl}
\SetMathAlphabet{\mathsfit}{bold}{\encodingdefault}{\sfdefault}{bx}{n}

\usepackage{hyperref}
\usepackage{url}
\usepackage{multirow}
\usepackage{booktabs}
\usepackage{graphicx}

\title{MapSAM2: Adapting SAM2 for Automatic Segmentation of Historical Map Images and Time Series}

\newcommand{\authorname}[1]{\textbf{#1}} % Ensures all names are bold

\author{\authorname{Xue Xia}\textsuperscript{1}, 
  \authorname{Randall Balestriero}\textsuperscript{2}, 
  \authorname{Tao Zhang}\textsuperscript{3}, \\
  \authorname{Yixin Zhou}\textsuperscript{1}, 
  \authorname{Andrew Ding}\textsuperscript{1}, 
  \authorname{Dev Saini}\textsuperscript{1},  
  \authorname{Lorenz Hurni}\textsuperscript{1} \\
  \textsuperscript{1}ETH Zurich, Zurich, Switzerland 
  \textsuperscript{2}Brown University, Providence, USA \\
  \textsuperscript{3}Wuhan University, Wuhan, China 
}

\iclrfinalcopy % Uncomment for camera-ready version, but NOT for submission.
\begin{document}

\maketitle

\begin{abstract}
Historical maps are unique and valuable archives that document geographic features across different time periods. However, automated analysis of historical map images remains a significant challenge due to their wide stylistic variability and the scarcity of annotated training data. Constructing linked spatio-temporal datasets from historical map time series is even more time-consuming and labor-intensive, as it requires synthesizing information from multiple maps. Such datasets are essential for applications such as dating buildings, analyzing the development of road networks and settlements, studying environmental changes etc. We present MapSAM2, a unified framework for automatically segmenting both historical map images and time series. Built on a visual foundation model, MapSAM2 adapts to diverse segmentation tasks with few-shot fine-tuning. Our key innovation is to treat both historical map images and time series as videos. For images, we process a set of tiles as a video, enabling the memory attention mechanism to incorporate contextual cues from similar tiles, leading to improved geometric accuracy, particularly for areal features. For time series, we introduce the annotated Siegfried Building Time Series Dataset and, to reduce annotation costs, propose generating pseudo time series from single-year maps by simulating common temporal transformations. Experimental results show that MapSAM2 learns temporal associations effectively and can accurately segment and link buildings in time series under limited supervision or using pseudo videos. We will release both our dataset and code to support future research.

\end{abstract}

\section{Introduction}
Historical maps offer valuable information for studying past landscapes and analyzing how territories, environments, and human settlements have evolved over time \citep{sun2021}. They serve as crucial resources across various scientific domains, including ecology, urban planning, archaeology, and environmental science \citep{heitzler2019, xia2024vectorizing}. The broad utility of the geographic information encoded in historical maps makes automatic segmentation a critical task. Moreover, many applications related to temporal change require not only the analysis of individual maps, but also the synthesis of information across entire historical map time series \citep{rath2025, harisena2025}.

Mainstream approaches primarily focus on the automatic segmentation of individual historical map images using deep learning models such as Convolutional Neural Networks (CNNs) or Vision Transformers \citep{heitzler2020, jiao2022, xia2023contrastive, lin2024hyper}. Segmenting historical map time series is typically handled through a multi-step pipeline built upon image-level segmentation: geographic features are first extracted from each map, followed by the alignment of corresponding entities across different years using heuristic methods, such as spatial distance or topological relations \citep{sun2021, shbita2020building}. In this context, the term alignment follows the definition in \citep{sun2021}, referring to the task of linking entities that represent the same real-world geographic object across time. However, this multi-step approach suffers from low automation and depends heavily on handcrafted linking rules, which are prone to failure under the varying distortions commonly found in historical maps across locations and time periods.

\begin{figure}[t]
  \centering
  \includegraphics[width=1\textwidth]{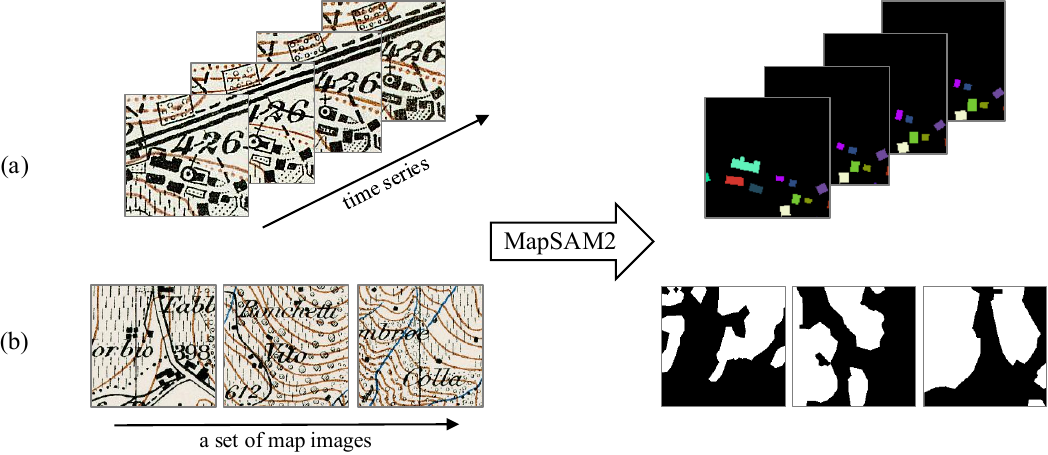}
  \caption{Segmentation capabilities of MapSAM2. MapSAM2 supports (a) instance-level segmentation and linking for historical map time series and (b) semantic segmentation for historical map images.}
  \label{fig:intro}
\end{figure}

In this paper, we introduce MapSAM2, the first framework capable of handling both historical map images and time series (Figure \ref{fig:intro}). For images, we focus on semantic segmentation, which is the most common task in historical map analysis. For time series, we address instance-level segmentation and linking across different years. MapSAM2 builds on recent advancements in the visual foundation model SAM2 \citep{ravi2024sam2}, which extends the original success of SAM \citep{kirillov2023} from zero-shot image segmentation to videos. Given user-provided prompts in the form of points, boxes, or masks on any frame to define the object of interest, SAM2 predicts a spatio-temporal mask for that object across the entire video. A key feature of SAM2 is its memory mechanism, which facilitates the sharing of feature embeddings across frames, allowing SAM2 to propagate mask predictions throughout the sequence. When applied to images, SAM2 treats each image as a single-frame video. In this case, the memory remains empty and the memory mechanism is not activated.

To adapt SAM2 to the historical map domain, MapSAM2 treats time series data as videos, allowing it to directly leverage SAM2’s strong generalization capabilities for video segmentation. For historical map images, in contrast to SAM2's original design, MapSAM2 introduces a novel perspective by treating a collection of images as a video. This allows the memory mechanism to be extended to image-based applications, where each input image can attend to previously seen, similar images to incorporate additional contextual cues. To make the system practical for processing tens of thousands of historical maps, we eliminate user interaction entirely. For semantic segmentation of map images, we find that training the default query tokens embedded in the mask decoder is sufficient to produce accurate masks, removing the need for explicit prompts. For time series data, where instance-level segmentation is required, we integrate a YOLO detector \citep{khanam2024yolov11} to automatically generate prompts for MapSAM2.

In addition, due to the domain gap between historical maps and natural images, model adaptation is necessary. We adopt Low-Rank Adaptation (LoRA) \citep{hu2022lora} to fine-tune the model efficiently with minimal computational overhead. Finally, given the scarcity and annotation cost of video-format training data, we propose a method for generating pseudo historical map time series by applying controlled transformations to single-year map images. This strategy enables effective fine-tuning for time series segmentation using only image-level annotations, significantly enhancing the practicality of applying video-based methods to historical map data.
 
We evaluate the performance of MapSAM2 on established benchmarks for historical map image segmentation, including tasks such as railway, vineyard, and building block detection \citep{xia2025mapsam, chazalon2021icdar}. MapSAM2 outperforms current state-of-the-art methods, particularly in the segmentation of areal features. The incorporation of memory attention further enhances performance compared to variants without it. 

To support time series segmentation, we curated the Siegfried Building Time Series Dataset, consisting of over 2,000 videos, each containing maps from four historical timestamps. This is the first video segmentation dataset in the historical map domain, and we make it publicly available to support future research. Additionally, we generate pseudo videos from single-year historical map images and release them alongside the real video dataset. Experimental results show that MapSAM2 can effectively segment and link historical map time series, even under few-shot training conditions and when using pseudo videos. We also release our code to promote transparency and reproducibility.

\section{Related work}

\subsection{Segment anything model family}
SAM \citep{kirillov2023} marks a significant advancement in visual foundation models by enabling promptable image segmentation, capable of producing high-quality object masks from simple prompts such as points, boxes, or masks. SAM2 \citep{ravi2024sam2} further extends this capability to the video domain by introducing a memory mechanism that captures relationships across frames. 

Trained on large-scale datasets, SAMs exhibit strong generalization capabilities and have been adopted across a wide range of applications. Research efforts have primarily focused on two directions: addressing SAMs' current limitations and extending their applicability to specialized domains. In the first direction, researchers have worked to overcome SAMs' prompt-dependent nature, which limits automation. This includes the development of prompt generation modules \citep{chen2024rsprompter, chen2024unsam} and methods that allow customisation from one-shot example \citep{zhang2023personalize, mao2025one}. Other work improves the quality of output masks, particularly in fine-grained segmentation tasks \citep{ke2023segment, shen2025mgd}. In the second direction, efforts have been made to adapt SAMs to domain-specific tasks such as medical imaging \citep{zhang2023customized, chen2024sam2, na2024segment} and remote sensing \citep{yan2023ringmo, ding2024adapting, li2025polyfootnet}. These adaptations often involve the use of lightweight tuning methods such as Adapters \citep{houlsby2019parameter} or LoRA \citep{hu2022lora} to inject domain-specific knowledge, along with targeted modules to further enhance performance.

Since the performance of SAMs in historical map segmentation, particularly for temporal localization in time series, remains largely underexplored and unproven, this paper aims to develop an automated pipeline for adapting SAM2 to segment both historical map images and time series.

\subsection{Historical map segmentation}
Most studies on historical map segmentation focus on semantic segmentation of individual map images. For example, \citet{xia2022cnn} apply CNN-based template matching to segment wetlands; \citet{heitzler2020} use U-Net for building footprint segmentation; \citet{lin2024hyper} apply deformable Transformers for text detection and recognition; and \citet{xia2023contrastive} employ Swin-Unet with contrastive pretraining for railway segmentation. To reduce annotation requirements, \citet{xia2025mapsam} propose MapSAM, which leverages the powerful, general-purpose feature representations of SAM for few-shot segmentation. MapSAM introduces several adaptations, including DoRA (Weight-Decomposed Low-Rank Adaptation) \citep{liu2024dora} in the image encoder for domain-specific adaptation, an auto-prompt generation module using a specialized CNN, and enhanced positional-semantic prompts with a masked-attention mask decoder.

In contrast, MapSAM2 benefits from the enhanced architecture of SAM2, which employs a hierarchical image encoder, Hiera \citep{ryali2023}, in place of SAM’s Vision Transformer (ViT) \citep{dosovitskiy2020}. While ViT maintains a uniform spatial resolution throughout, Hiera’s hierarchical design produces multiscale features, enabling skip connections that enrich the mask decoder with multi-scale context. These architectural advantages allow MapSAM2 to simplify the overall design: we retain SAM2’s original encoder-decoder structure, introducing only LoRA and the memory mechanism to process sets of images as pseudo-videos. No additional specialized modules are used, reflecting Occam’s Razor: when a simpler method achieves superior results, added complexity is not only unnecessary but potentially counterproductive \citep{sterzinger2025few}.

The temporal dimension is also a critical characteristic of historical maps. Many important socioeconomic and ecological studies require consistent time-series segmentation that goes beyond individual images \citep{rath2025, harisena2025}. However, historical map time series segmentation remains largely underexplored. Existing approaches typically rely on post-processing steps to associate features across time, using spatial distance or topological relations \citep{sun2021}. \citet{xia2024video} are the first to propose an end-to-end approach using the video segmentation model Mask2Former-VIS \citep{cheng2021mask2former}, which outperforms traditional two-step pipelines that combine Mask R-CNN \citep{he2017mask} with topological linking \citep{clementini1997}. MapSAM2 builds on this video-based paradigm but goes further by leveraging a foundation model for video segmentation, reducing the need for labor-intensive annotations and making the approach more practical and scalable for real-world historical map analysis.

\section{Method}
\begin{figure}[t]
  \centering
  \includegraphics[width=1\textwidth]{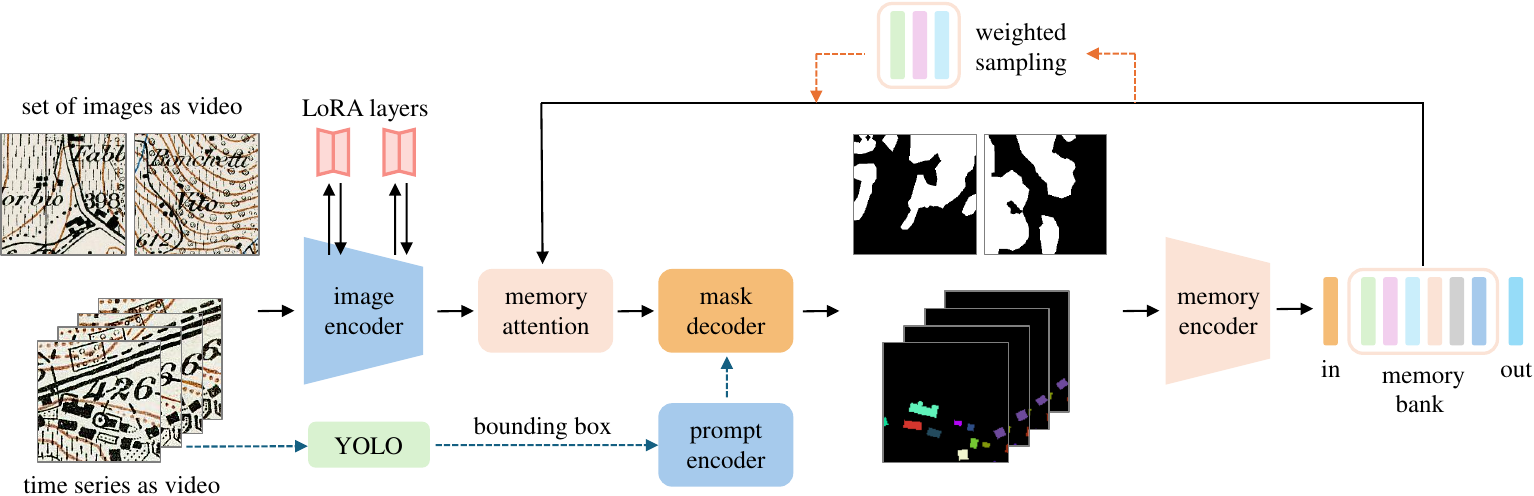}
  \caption{The MapSAM2 architecture. We propose treating both historical map time series and sets of map images as videos to enable memory-enhanced historical map segmentation. For time series data, YOLO is used to provide bounding box prompts, and a first-in–first-out strategy is applied to build the memory bank using the \(k\) most recent frames for memory attention. For images, no external prompts are provided; instead, the memory bank is constructed based on confidence and dissimilarity, followed by weighted sampling to select the \(k\) most relevant frames for memory attention. In the figure, solid arrows indicate operations common to both data types, blue dashed arrows denote operations specific to time series, and orange dashed arrows denote operations specific to images.}
  \label{fig:framework}
\end{figure}

Figure \ref{fig:framework} illustrates the overall framework of MapSAM2. Built upon the SAM2 architecture, it includes a LoRA-adapted image encoder, a prompt encoder, a mask decoder, and memory components such as the memory encoder, memory bank, and memory attention module. When processing historical map images, we eliminate the use of external prompts and instead fine-tune the mask decoder to generate masks directly. For historical map time series, we integrate a YOLO detector \citep{redmon2016you} to provide automatic prompts. Further details are presented in the following section.

\subsection{LoRA-adapted image encoder}
For out-of-distribution data such as historical maps, which exhibit domain-specific characteristics and differ fundamentally from the natural images SAM2 was originally trained on, fine-tuning is necessary but also computationally expensive. Full fine-tuning can lead to the forgetting of pre-trained features and may degrade the model’s generalization ability \citep{marti2025}. To mitigate domain discrepancy, preserve generalization, while keeping computational costs low, we adopt Low-Rank Adaptation (LoRA) to efficiently fine-tune the image encoder of SAM2.

More specifically, we freeze the pre-trained weight matrix \( W \in \mathbb{R}^{d \times k} \) in the SAM2 image encoder and compute the weight update \( \Delta W \in \mathbb{R}^{d \times k} \) through a low-rank decomposition, expressed as \( \Delta W = BA \), where \( B \in \mathbb{R}^{d \times r} \) and \( A \in \mathbb{R}^{r \times k} \) are low-rank matrices. The rank \( r \) satisfies \( r \ll \min(d, k) \), which significantly reduces the number of trainable parameters. We apply this low-rank adaptation to both the query and value projection layers in each transformer block of the image encoder:
\begin{equation}
Q' = (W_q + B_q A_q) \cdot x, \quad V' = (W_v + B_v A_v) \cdot x
\end{equation}
where \( x \) represents the input image tokens, and \( Q' \) and \( V' \) are the projected queries and values. During fine-tuning, the pre-trained weights \( W_q \) and \( W_v \) are kept frozen, while the low-rank matrices \( B_q \), \( A_q \), \( B_v \), and \( A_v \) serve as a trainable bypass to achieve the weight update.

\subsection{Segment historical map images as a video}
When historical maps lack temporal continuity, or when modeling temporal associations is unnecessary, they are commonly divided into a complete set of smaller map tiles for segmentation. This collection can be treated as a single, extended pseudo-video sequence, processed in a streaming manner. Frames are ingested sequentially and encoded into memory for use by subsequent frames. We leverage a memory mechanism to condition the embeddings of the current tile on those of similar tiles stored in a memory bank. To construct a diverse and high-quality memory set, we adopt the self-sorting memory bank proposed in MedSAM-2 \citep{zhu2024}, which dynamically updates the memory bank and selects the most informative embeddings for memory attention. This approach is more effective than simply using the most recent \( k \) frames as in SAM2, since the input is not a real temporal video and the notion of "recent" is not meaningful in this context.

\textbf{Self-sorting memory bank.} The self-sorting memory bank consists of two main components. First, it dynamically updates the memory bank based on the confidence and dissimilarity of candidate embeddings, ensuring the memory bank remains diverse so that each incoming tile can retrieve relevant information. Second, it selects the most relevant embeddings from the memory bank to compute memory attention with the incoming embedding. 

More specifically, given a candidate embedding \( E_t \), the model predicts the segmentation mask \( y_t \) and computes the IoU confidence score \( c_t \) using the mask decoder. If the confidence score exceeds a predefined threshold, \( E_t \) is considered for inclusion in the memory bank \( \mathcal{M}_{t-1} \) based on dissimilarity. This is done by forming a candidate set of memory embeddings \( \mathcal{C} = \mathcal{M}_{t-1} \cup \{ E_t \} \), and then selecting the top \( K \) embeddings with the highest total dissimilarity \( D_i \) to form the updated memory bank \( \mathcal{M}_t \):

\begin{equation}
D_i = \sum_{\substack{E_j \in \mathcal{C} \\ j \ne i}} \left( 1 - \text{sim}(E_i, E_j) \right), \quad \forall E_i \in \mathcal{C},
\end{equation}

\begin{equation}
\mathcal{M}_t = \text{TopK}_{E_i \in \mathcal{C}} (D_i),
\end{equation}

where \( K \) is the memory bank size, and \( \text{sim}(\cdot, \cdot) \) denotes the cosine similarity function.

For the next incoming tile \( F_{t+1} \), before it interacts with the updated memory bank \( \mathcal{M}_t \), we resample the memory bank to select the \( k \) most similar embeddings to \( F_{t+1} \), based on the following probability distribution \( \{p_{i,t}\} \):

\begin{equation}
p_{i,t} = \frac{\operatorname{sim}(F_{t+1}, E_i)}{\sum\limits_{E_j \in \mathcal{M}_t} \operatorname{sim}(F_{t+1}, E_j)}, \quad \forall E_i \in \mathcal{M}_t.
\end{equation}

Higher selection probabilities are thus assigned to embeddings that are more similar to \( F_{t+1} \), enhancing the relevance of the memory bank when computing memory attention.

\textbf{Image segmentation without prompts.} The mask decoder processes the frame embeddings conditioned on the self-sorting memory bank to produce a prediction. Unlike in SAM2, we do not provide any additional prompts to the decoder. Instead, we leverage the default query tokens inherent in the mask decoder. By initializing the decoder with pretrained SAM2 parameters and allowing it to be trainable during fine-tuning, the model can perform automatic segmentation without requiring manual prompts.

\subsection{Segment historical map time series as videos}
\label{sec:3D branch}
With the addition of the temporal dimension, a time series of maps forms a 3D spatio-temporal volume that can naturally be treated as a video. This format more closely resembles natural videos, on which SAM2 was trained, and can therefore be processed using the same memory mechanism as in SAM2. Given an input time series of maps \( X = \{x_t\}_{t=1}^T \), we first extract a feature embedding \( F_t \) for each frame \( x_t \) using the LoRA-adapted image encoder. The memory attention mechanism conditions the current frame embedding \( F_t \) on past frame features stored in a memory bank via self-attention and cross-attention, resulting in a fused visual embedding \( E_t \). We use YOLOv11 \citep{khanam2024yolov11} to automatically generate bounding box prompts for each instance in \( x_t \), which are then transformed into embeddings \( P_t \) by the prompt encoder. The mask decoder takes \( E_t \) and \( P_t \) as input to produce the corresponding prediction mask \( M_t \). In cases where no prompt is provided for a frame, object information is propagated across frames through memory attention, enabling the mask decoder to generate segmentation masks solely based on context. Finally, the predicted mask \( M_t \) is passed through the memory encoder. Its output is summed with the unconditioned frame embedding \( F_t \) to produce memory features, which are then stored in the memory bank for computing memory attention in subsequent frames.

\textbf{Video segmentation with YOLO-based prompts.} Since we are performing instance-level segmentation and linking on historical map time series, prompting is essential to distinguish individual objects. To this end, we employ a pre-trained YOLOv11 and fine-tune it on historical map data. Each map in the time series is processed individually by YOLO, which detects objects by extracting multi-scale visual features with CNNs and outputs bounding boxes. These bounding boxes serve as input prompts, defining the objects of interest for which spatio-temporal masks are predicted.

\begin{figure}[!b]
  \centering
  \includegraphics[width=0.6\textwidth]{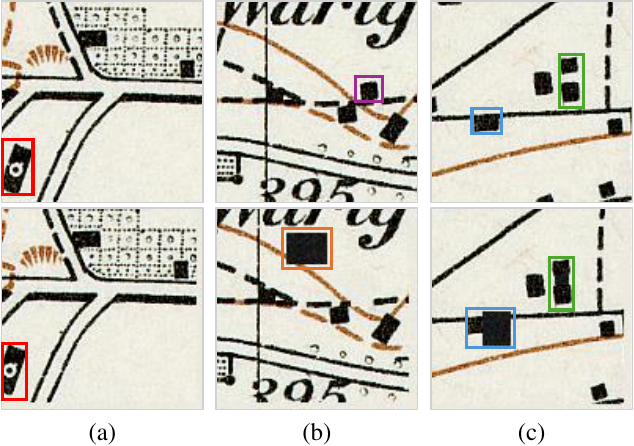}
  \caption{Generating pseudo time series by transforming single-year maps. The applied transformations are highlighted with bounding boxes in the examples: (a) shift, (b) appearance and disappearance, and (c) shape change and merge.}
  \label{fig:transform}
\end{figure}

\textbf{Learning from sparse annotation.} A key challenge in applying video segmentation models to historical map time series is the lack of video-format annotations, which must provide not only object masks but also association information across frames. Since obtaining such annotations is prohibitively expensive, while image-level annotations are more commonly available for historical maps, we propose constructing pseudo videos from these image data. This approach enables the training of video segmentation models when only sparse image-level annotations are available. 

Pseudo videos are created to approximate real map time series by mimicking the major transformations observed across different years, such as object shifts, appearances, disappearances, and merges. We apply random combinations of these transformations to a source map and its associated mask to create an annotated two-frame pseudo video. Prior research has shown that two-frame videos are sufficient for training video segmentation models \citep{wang2024}. Therefore, we limit our synthetic sequences to two frames to reduce complexity and avoid the ambiguities that can arise in longer sequences.

As illustrated in Figure~\ref{fig:transform}, the \textbf{shift} transformation simulates distortions in historical maps by randomly shifting the image along the $x$ and $y$ axes by $\pm 5$ pixels. The \textbf{appearance} transformation mimics the emergence of new objects, while \textbf{disappearance} simulates the removal of existing ones. Since our target objects are buildings, which are symbolized as square constructions composed of black pixels, we simulate the appearance of a new building by inserting a black rectangle with a random height and width ranging from 5 to 30 pixels. Disappearing buildings are removed by filling their regions with background pixels. If a newly added building overlaps with an existing one, this is treated as a \textbf{shape change}, and the overlapping region inherits the original building's instance ID in the mask. In contrast, newly appeared buildings with no overlap are assigned new instance IDs. The \textbf{merge} transformation simulates cases where several small buildings are later combined into a larger structure. To indicate their association, both the original and merged buildings share the same instance ID. This is achieved by identifying the closest neighboring objects in the mask, dilating their masks to connect them, and then eroding the merged region to remove excess pixels while preserving the merging effect. The corresponding areas in the map image are filled with building pixels according to the new merged structure.

\subsection{Training strategy}
During fine-tuning, the image encoder and prompt encoder are frozen, while the LoRA layers, memory module, and mask decoder remain trainable. For semantic segmentation of historical map images, the model predicts a binary mask supervised using binary cross-entropy loss. For instance segmentation and linking in historical map time series, the model predicts a binary mask for each object instance, also supervised with binary cross-entropy loss. The overall loss is computed by summing and averaging the individual object losses across the entire video sequence.

\section{Experiment}

\subsection{Datasets}
Historical map image segmentation is evaluated on the same datasets used in MapSAM \citep{xia2025mapsam}, namely the Siegfried Railway and Vineyard Dataset and the ICDAR 2021 Building Block Dataset \citep{chazalon2021icdar}. Historical map time series segmentation is evaluated on the \textbf{Siegfried Building Time Series Dataset}, which we curated and publicly released to support community research and development. It is derived from the Swiss Siegfried maps (\copyright~swisstopo) spanning four timestamps, namely year 1896, 1904, 1932, and 1945. Buildings from these maps were manually digitized and assigned instance IDs. Linked building instances across years share the same ID to indicate correspondence, while instance uniqueness is preserved within each individual frame. 

Each map tile is sized at $128 \times 128$ pixels, and tiles from the same geographic location but different years are grouped to form video-format inputs. The resulting dataset contains 2,105 training videos, 283 validation videos, and 326 test videos, with each video consisting of four frames.

In addition to real time series, we also generate \textbf{pseudo videos} from single-year historical map images. Specifically, we select one frame from each real video (e.g., from the year 1945) and apply the transformations described in Section~\ref{sec:3D branch} to simulate temporal changes, resulting in a two-frame pseudo video. This process is applied to both the training and validation sets, ensuring that the number of pseudo training and validation videos matches that of the original sets.

To evaluate model performance in low-data regimes, we randomly sample 10 videos each from the training and validation splits of both the real and pseudo datasets. The test set remains the same across all evaluation scenarios.

\subsection{Evaluation metrics}
We use Intersection over Union (IoU) to evaluate image segmentation performance. For time series segmentation, we report precision, recall, and F1-score instead of metrics such as J\&F (used in SAM2) or AP (used in Mask2Former-VIS). This unified evaluation protocol enables fair comparison across a diverse set of methods, including foundation models (MapSAM2), standard video instance segmentation models (Mask2Former-VIS), and traditional two-step pipelines (Mask R-CNN followed by instance linking). This choice is particularly important because traditional pipelines do not produce confidence scores, making metrics like AP inapplicable.

Concretely, each linked instance $\{ e_1^i, e_2^i, e_3^i, e_4^i \}$ corresponds to the same object across four timestamps, with empty masks used for frames where the object does not appear. A predicted instance is considered a true positive (TP) if its IoU with a ground truth instance exceeds 0.5. Instances without a matching ground truth are counted as false positives (FP), while undetected ground truth instances are treated as false negatives (FN). Unlike in image instance segmentation, a video instance is a sequence of masks. Therefore, the IoU is computed both spatially and temporally by summing the intersections and unions over all frames:
\begin{equation}
\text{IoU}(i, j) =
\frac{\sum_{t=1}^{4} \left| e_t^i \cap g_t^j \right|}
{\sum_{t=1}^{4} \left| e_t^i \cup g_t^j \right|}
\end{equation}
where $g_t^j$ denotes the ground truth. Based on these TP, FP, and FN counts, we compute precision, recall, and F1-score accordingly.

\subsection{Implementation}
We conduct all experiments using the \texttt{sam2\_hiera\_small} version of SAM2 as the backbone, loading its pretrained weights accordingly. All experiments are implemented on a single NVIDIA Quadro RTX 5000 GPU with 16 GB of memory. We use the AdamW optimizer with a learning rate of $1 \times 10^{-4}$, weight decay of $1 \times 10^{-4}$, and train for 200 epochs, retaining the model with the best validation accuracy. The LoRA adaptation uses a rank of 4 ($r=4$). For image segmentation, we use a batch size of 2; for time series segmentation, the batch size is set to 1 due to higher memory requirements. In the time series setting, we reverse the chronological order of frames so that the latest year appears first. During training, we randomly sample two frames from the full four-frame video sequences. Prompts are provided only for the first frame, both during training and testing.

\subsection{Historical map images}

\subsubsection{Main result}

\begin{table*}[b]
    \centering
    \caption{Image segmentation accuracy (IoU) on the Siegfried Railway, Vineyard, and ICDAR 2021 Building Block datasets under full and few-shot settings.}
    \vspace{1\baselineskip}
    \label{tab:2d-results}
    \small
    \begin{tabular}{l c c c c c c c}
        \toprule
        \multirow{2}{*}{\textbf{Method}} 
        & \multicolumn{4}{c}{\textbf{Railway (5872)}} 
        & \multicolumn{2}{c}{\textbf{Vineyard (613)}} 
        & \multicolumn{1}{c}{\textbf{Building Block}} \\
        & Full & 10\% & 1\% & 10-shot 
        & Full & 10-shot 
        & 10-shot \\
        \midrule

        U-Net \citep{ronneberger2015u}           & \textbf{91.9} & \textbf{90.6} & 83.5 & 61.4 & 77.0 & 60.2 & 60.0 \\
        PerSAM \citep{zhang2023personalize}         & --            & --            & --   & 5.9  & --   & 22.7 & 16.0 \\
        Few-Shot SAM \citep{wu2023self}    & --            & --            & --   & 35.8 & --   & 46.8 & 15.5 \\
        SAMed \citep{zhang2023customized}           & 86.3          & 85.7          & 86.0 & 75.4 & 74.9 & 61.5 & 70.3 \\
        MapSAM \citep{xia2025mapsam} & 89.5          & 88.7          & \textbf{86.5} & \textbf{78.5} 
                        & 74.3 & 60.0 & 71.1 \\
        MapSAM2 (Ours)        & 90.9          & 89.8          & 84.7 & 73.0 & \textbf{77.3} & \textbf{67.6} & \textbf{75.8} \\
        \bottomrule
    \end{tabular}
\end{table*}

We evaluate MapSAM2 for historical map image segmentation across maps of varying cartographic styles and feature types, including both linear and areal geographic entities. Results are presented in Table~\ref{tab:2d-results}. MapSAM2 achieves the best performance on areal features, such as vineyards and building blocks, under both full and few-shot training regimes. Notably, while other fine-tuned foundation model baselines, such as MapSAM \citep{xia2025mapsam} and SAMed \citep{zhang2023customized}, fail to surpass the domain-specific U-Net when sufficient training data is available, MapSAM2 does. For example, on the vineyard dataset with full training, MapSAM2 achieves an IoU of 77.3, marginally outperforming U-Net (77.0), and significantly outperforming MapSAM by 3\% in the full setting and by 7.6\% in the 10-shot setting. These results highlight the effectiveness of MapSAM2 in adapting foundation models to the domain of historical maps.

However, on linear features such as railways, MapSAM2 shows more modest performance. While it achieves slight improvements over MapSAM under full-data and 10\% training conditions (by approximately 1\%), it underperforms MapSAM in low-data scenarios (1\% and 10-shot). Recent findings \citep{raghu2021, wang2022fourier} suggest that attention mechanisms function as low-pass filters, emphasizing low-frequency information and global context while suppressing high-frequency details. This may explain why MapSAM2, through its introduced memory attention mechanism, achieves greater performance gains on broad areal features such as vineyards and building blocks, but is less effective on narrow, high-frequency structures such as railways.

\subsubsection{Effectiveness of memory attention}
We conduct an ablation study to evaluate the effectiveness of memory attention in MapSAM2. As shown in Table~\ref{tab:ma}, removing memory attention leads to a noticeable performance drop for both linear and areal features. For instance, under the 10-shot training setting, incorporating memory attention improves the segmentation accuracy by 16.1\% IoU for railways and 14.3\% for vineyards. The performance gain is more pronounced in few-shot settings compared to those with sufficient training data. This demonstrates that memory attention significantly enhances MapSAM2’s ability to leverage additional contextual cues, leading to improved segmentation accuracy for both linear and areal features.

\begin{figure}[t]
  \centering
  \includegraphics[width=0.8\textwidth]{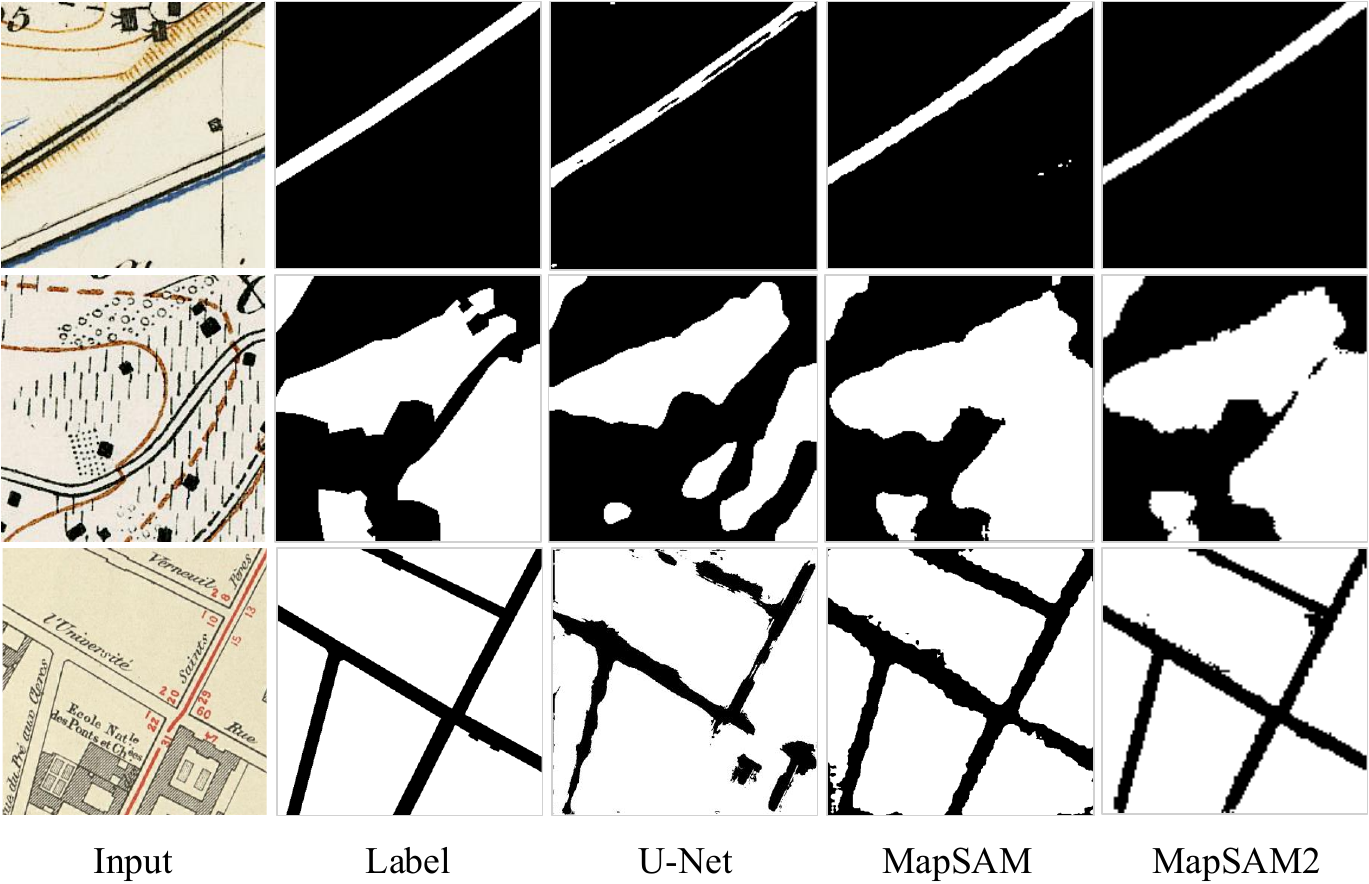}
  \caption{Image segmentation results from U-Net, MapSAM, and MapSAM2, each trained with 10-shot samples for detecting railway, vineyard, and building block.}
  \label{fig:2dresult}
\end{figure}

\begin{table*}[t]
    \centering
    \caption{Ablation study on the effectiveness of memory attention.}
    \vspace{1\baselineskip}
    \label{tab:ma}
    \small
    \begin{tabular}{c c c c c}
        \toprule
        \multirow{2}{*}{\textbf{Memory Attention}} 
        & \multicolumn{2}{c}{\textbf{Railway}}
        & \multicolumn{2}{c}{\textbf{Vineyard}} \\
        & 10\% & 10-shot 
        & Full & 10-shot \\
        \midrule

        w/        & \textbf{89.8}    & \textbf{73.0}    & \textbf{77.3} & \textbf{67.6} \\ 
        w/o       & 85.4       & 56.7    & 72.0          & 53.3 \\
        \bottomrule
    \end{tabular}
\end{table*}

\subsubsection{Visualization}
Figure \ref{fig:2dresult} presents a visual comparison of image segmentation results obtained by MapSAM2 and other baseline models under the 10-shot training setting. As shown in the figure, MapSAM2 produces clearer boundaries at edge pixels and more accurate geometric shapes compared to MapSAM and U-Net, particularly for areal features such as vineyards and building blocks. These results highlight MapSAM2’s strong capability for efficient segmentation of historical map images in low-data scenarios.

\subsection{Historical map time series}

\subsubsection{Main result}
We compare the performance of MapSAM2 in segmenting and linking building instances from historical map time series with two baseline models, Mask2Former-VIS \citep{cheng2021mask2former} and Mask R-CNN \citep{he2017mask} combined with post-hoc linking. All models are evaluated on the same test set, with training conducted on both real and pseudo video datasets under full (2,105 videos) and limited (10-shot) supervision. The experimental results are reported in Table \ref{tab:3d-results}.

When trained on the real building time series dataset, Mask2Former-VIS achieves the highest performance under full supervision. However, its performance drops sharply in the 10-shot setting. In contrast, MapSAM2 demonstrates robust performance under limited supervision, outperforming Mask2Former-VIS and Mask R-CNN with linking by 35.8\% and 15.7\% in F1 score, respectively. The pseudo building time series dataset, generated by transforming the labeled image dataset, provides a promising alternative when real video-format annotations are unavailable. MapSAM2 achieves an F1 score of 83.1 on the full pseudo dataset and 71.1 in the 10-shot setting, comparable to results obtained on the real time series dataset.

\begin{table*}[t]
    \centering
    \caption{Video segmentation accuracy on real and pseudo Siegfried building time series datasets under full (2105 videos) and 10-shot training settings. The best performance under each training setting is highlighted in \textbf{bold}.}
    \vspace{1\baselineskip}
    \label{tab:3d-results}
    \small
    \begin{tabular}{ll|ccc|ccc}
        \toprule
        \textbf{Dataset} & \textbf{Method} 
        & \multicolumn{3}{c|}{\textbf{Full Training}} 
        & \multicolumn{3}{c}{\textbf{10-shot Training}} \\
        &  
        & Prec. & Rec. & F1 
        & Prec. & Rec. & F1 \\
        \midrule

        \multirow{3}{*}{Real} 
        & Mask R-CNN+Link    
        & 72.9 & 60.9 & 66.4 
        & 62.5 & 48.9 & 54.9 \\
        & Mask2Former-VIS    
        & \textbf{92.9} & \textbf{86.4} & \textbf{89.5} 
        & \textbf{76.4} & 22.5 & 34.8 \\
        & MapSAM2             
        & 85.6 & 82.2 & 83.9 
        & 73.8 & \textbf{67.6} & \textbf{70.6} \\

        \midrule

        \multirow{3}{*}{Pseudo} 
        & Mask R-CNN+Link    
        & 73.7 & 63.9 & 68.4 
        & 57.0 & 51.3 & 54.0 \\
        & Mask2Former-VIS    
        & 84.7 & 77.6 & 81.0 
        & \textbf{75.7} & 42.1 & 54.1 \\
        & MapSAM2             
        & \textbf{84.8} & \textbf{81.5} & \textbf{83.1} 
        & 72.5 & \textbf{69.7} & \textbf{71.1} \\

        \bottomrule
    \end{tabular}
\end{table*}

\begin{table*}[t]
    \centering
    \caption{Effectiveness of Prompt Quality. We fix MapSAM2 to the 10-shot video training setting (on both real and pseudo datasets) and vary only the data used to train the YOLO detector from 10-shot to the full dataset to assess the impact of prompt quality on segmentation performance.}
    \vspace{1\baselineskip}
    \label{tab:prompt}
    \small
    \begin{tabular}{c c c c c c c}
        \toprule
        \multirow{2}{*}{\textbf{YOLO Training}} 
        & \multicolumn{3}{c}{\textbf{Real}}
        & \multicolumn{3}{c}{\textbf{Pseudo}} \\
        & Prec. & Rec. & F1
        & Prec. & Rec. & F1 \\
        \midrule
        
        10-shot     & 73.8          & 67.6          & 70.6  
                    & 72.5          & 69.7          & 71.1 \\
        Full        & \textbf{85.1} & \textbf{81.7} & \textbf{83.4} 
                    & \textbf{84.2} & \textbf{80.9} & \textbf{82.5} \\ 
        \bottomrule
    \end{tabular}
\end{table*}

\subsubsection{Effectiveness of prompt quality}
Since image-level annotations are more commonly available than video-level annotations for historical maps, a practical approach for processing historical map time series is to fine-tune the video segmentation model on a small number of annotated videos to inject temporal domain knowledge, while leveraging a larger set of image-level annotations to train a high-quality YOLO detector for prompt generation. Given that the video segmentation backbone is based on a vision foundation model, low-resource fine-tuning can still yield strong results, and improved prompt quality can further boost performance.

In Table~\ref{tab:3d-results}, we report results using the same dataset to train both the YOLO detector and MapSAM2. In contrast, Table~\ref{tab:prompt} isolates the effect of prompt quality by fixing MapSAM2 to the 10-shot video training setting (on both real and pseudo datasets) while varying only the data used to train the YOLO detector, from 10-shot to the full dataset. The results show that higher-quality prompts, produced by a better-trained detector, can significantly enhance MapSAM2's performance. For example, using YOLO trained on the full real building time series for prompting improves MapSAM2's F1 score by 12.8\% compared to using YOLO trained with only 10-shot data. This demonstrates that low-resource fine-tuning of MapSAM2, when combined with high-quality prompts, can achieve the strongest performance with minimal annotation effort.

\begin{figure}[t]
  \centering
  \includegraphics[width=1\textwidth]{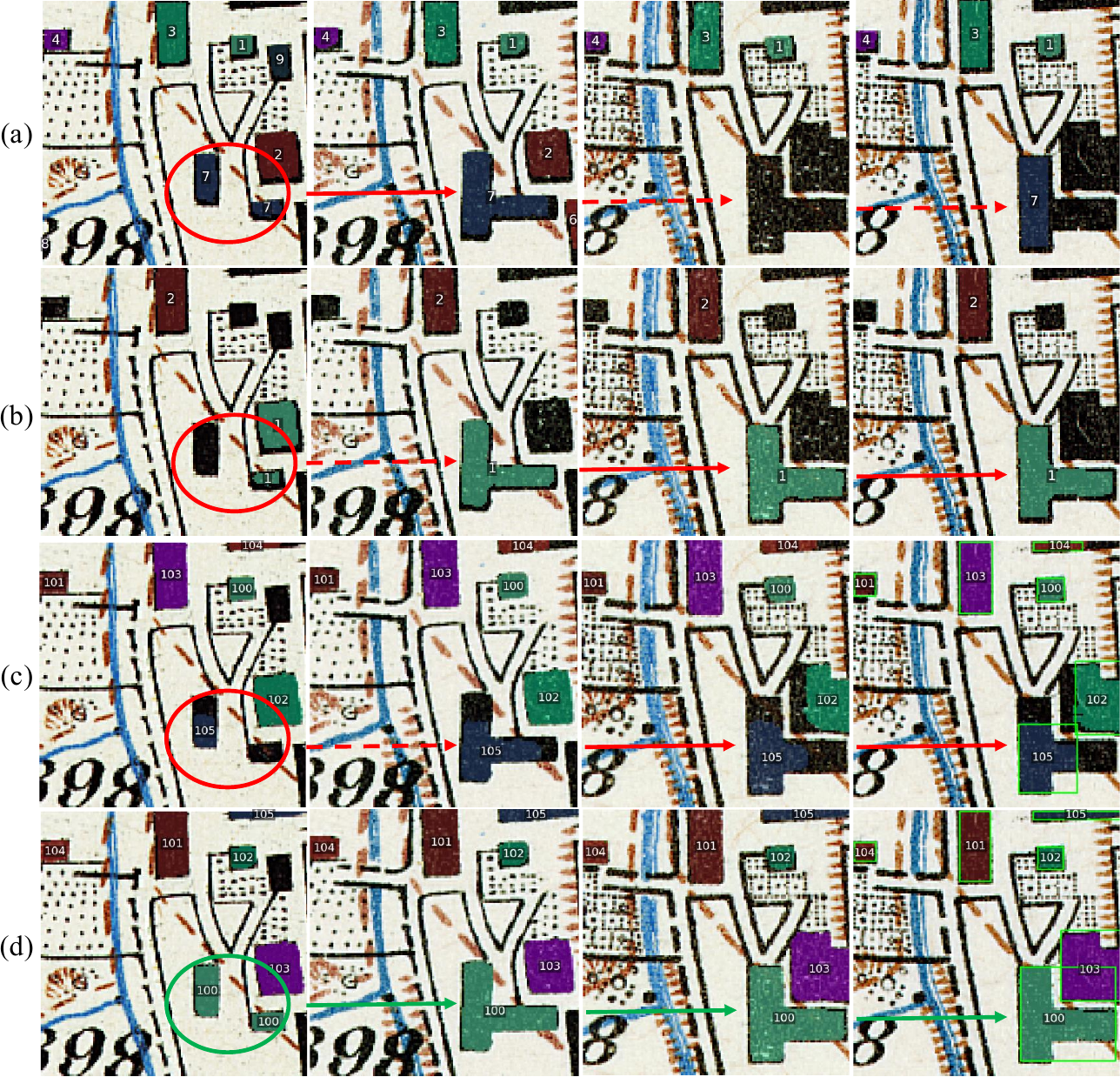}
  \caption{Video segmentation results under 10-shot training on the real Siegfried Building Time Series: (a) Mask R-CNN with linking, (b) Mask2Former-VIS, (c) MapSAM2 prompted by YOLO trained on the same 10-shot data, and (d) MapSAM2 prompted by YOLO trained on the full dataset. The YOLO prompt is provided only for the latest frame and is shown as a green bounding box. A challenging case, where two small buildings merge into a larger structure over time, is highlighted with a circle (green indicates successful video segmentation, red indicates failure). Links are indicated with arrows: solid arrows denote correct links, while dashed arrows denote incorrect links.}
  \label{fig:3dresult}
\end{figure}

\subsubsection{Visualization}
We visualize video segmentation results obtained by Mask R-CNN with linking, Mask2Former-VIS, and MapSAM2 under 10-shot training on the real Siegfried Building Time Series Dataset in Figure~\ref{fig:3dresult}. For MapSAM2, we further present results under two prompting conditions: one using a YOLO detector trained on the same 10-shot data and another trained on the full dataset. In both cases, the YOLO prompt is provided only for the latest frame and is shown as a green bounding box in the figure. Due to limited training samples, both Mask R-CNN with linking and Mask2Former-VIS exhibit numerous missed detections. In contrast, MapSAM2 successfully segments and tracks most instances, with performance further improved when higher-quality prompts are available.

A particularly challenging case involves two small buildings merging into a larger structure over time (circled in the figure). Only MapSAM2 with high-quality prompts correctly segments and links this complex merging scenario. With lower-quality prompts, MapSAM2 fails to capture the complete geometry of the merged building due to inaccurate bounding boxes, resulting in the loss of one small building in tracking. However, when the bounding box accurately covers the full structure, all components are successfully tracked. This demonstrates the strength of MapSAM2’s memory attention mechanism, which enables effective temporal communication across frames.

Although MapSAM2 with high-quality prompts achieves the best performance, we still observe a missing building in the first frame. This occurs because in our experimental setup, prompts are only provided for the latest frame. As a result, buildings that appear in earlier frames but not in the latest one lose their track, since no prompt is assigned to them. A potential solution would be to add prompts in additional frames, either interactively as in SAM2, or automatically by heuristically matching YOLO-detected bounding boxes across frames and extending the prompt set. However, to avoid leaking linkage information, we deliberately adopt the simple strategy of prompting only on the latest frame. Since the latest frame contains the majority of buildings, this approach performs well in most cases. Future research could explore more automatic and heuristic-free ways to address this limitation.

\section{Conclusion}
We present MapSAM2, an efficient adaptation of SAM2 for historical map image and time series segmentation. Our key innovation is to treat both historical map time series and sets of map images as videos, enabling memory-enhanced segmentation. Compared to MapSAM, MapSAM2 adopts a simpler yet more effective design, consisting of a LoRA-adapted image encoder, memory modules, and a mask decoder. This design achieves superior performance in segmenting historical maps of diverse styles, particularly for areal features. MapSAM2 is also highly effective for processing historical map time series, offering significant improvements in automation and accuracy over traditional multi-step pipelines, while substantially reducing annotation costs compared to standard video segmentation models. Furthermore, our pseudo-video generation strategy—transforming individual images to mimic common temporal changes in historical map time series—proves effective in training video segmentation models, providing a practical and scalable solution for historical map time series analysis.

\subsubsection*{Acknowledgments}
This research was funded by the Swiss National Science Foundation as part of the EMPHASES Project [Grant Number: 200021\_192018].

\bibliography{iclr2025_conference}

\begin{thebibliography}{46}
\providecommand{\natexlab}[1]{#1}
\providecommand{\url}[1]{\texttt{#1}}
\expandafter\ifx\csname urlstyle\endcsname\relax
  \providecommand{\doi}[1]{doi: #1}\else
  \providecommand{\doi}{doi: \begingroup \urlstyle{rm}\Url}\fi

\bibitem[Chazalon et~al.(2021)Chazalon, Carlinet, Chen, Perret, Dum{\'e}nieu, Mallet, G{\'e}raud, Nguyen, Nguyen, Baloun, et~al.]{chazalon2021icdar}
Joseph Chazalon, Edwin Carlinet, Yizi Chen, Julien Perret, Bertrand Dum{\'e}nieu, Cl{\'e}ment Mallet, Thierry G{\'e}raud, Vincent Nguyen, Nam Nguyen, Josef Baloun, et~al.
\newblock Icdar 2021 competition on historical map segmentation.
\newblock In \emph{International Conference on Document Analysis and Recognition}, pp.\  693--707. Springer, 2021.

\bibitem[Chen et~al.(2024{\natexlab{a}})Chen, Liu, Chen, Zhang, Li, Zou, and Shi]{chen2024rsprompter}
Keyan Chen, Chenyang Liu, Hao Chen, Haotian Zhang, Wenyuan Li, Zhengxia Zou, and Zhenwei Shi.
\newblock Rsprompter: Learning to prompt for remote sensing instance segmentation based on visual foundation model.
\newblock \emph{IEEE Transactions on Geoscience and Remote Sensing}, 62:\penalty0 1--17, 2024{\natexlab{a}}.

\bibitem[Chen et~al.(2024{\natexlab{b}})Chen, Lu, Zhu, Ding, Yu, Ji, Li, Sun, Mao, and Zang]{chen2024sam2}
Tianrun Chen, Ankang Lu, Lanyun Zhu, Chaotao Ding, Chunan Yu, Deyi Ji, Zejian Li, Lingyun Sun, Papa Mao, and Ying Zang.
\newblock Sam2-adapter: Evaluating \& adapting segment anything 2 in downstream tasks: Camouflage, shadow, medical image segmentation, and more.
\newblock \emph{arXiv preprint arXiv:2408.04579}, 2024{\natexlab{b}}.

\bibitem[Chen et~al.(2024{\natexlab{c}})Chen, Xu, Liu, and Yuan]{chen2024unsam}
Zhen Chen, Qing Xu, Xinyu Liu, and Yixuan Yuan.
\newblock Un-sam: Universal prompt-free segmentation for generalized nuclei images.
\newblock \emph{arXiv preprint arXiv:2402.16663}, 2024{\natexlab{c}}.

\bibitem[Cheng et~al.(2021)Cheng, Choudhuri, Misra, Kirillov, Girdhar, and Schwing]{cheng2021mask2former}
Bowen Cheng, Anwesa Choudhuri, Ishan Misra, Alexander Kirillov, Rohit Girdhar, and Alexander~G. Schwing.
\newblock Mask2former for video instance segmentation.
\newblock \emph{arXiv preprint arXiv:2112.10764}, 2021.
\newblock URL \url{https://arxiv.org/abs/2112.10764}.

\bibitem[Clementini \& Di~Felice(1997)Clementini and Di~Felice]{clementini1997}
Eliseo Clementini and Paolino Di~Felice.
\newblock Approximate topological relations.
\newblock \emph{International journal of approximate reasoning}, 16\penalty0 (2):\penalty0 173--204, 1997.

\bibitem[Ding et~al.(2024)Ding, Zhu, Peng, Tang, Yang, and Bruzzone]{ding2024adapting}
Lei Ding, Kun Zhu, Daifeng Peng, Hao Tang, Kuiwu Yang, and Lorenzo Bruzzone.
\newblock Adapting segment anything model for change detection in vhr remote sensing images.
\newblock \emph{IEEE Transactions on Geoscience and Remote Sensing}, 62:\penalty0 1--11, 2024.

\bibitem[Dosovitskiy et~al.(2020)Dosovitskiy, Beyer, Kolesnikov, Weissenborn, Zhai, Unterthiner, Dehghani, Minderer, Heigold, Gelly, et~al.]{dosovitskiy2020}
Alexey Dosovitskiy, Lucas Beyer, Alexander Kolesnikov, Dirk Weissenborn, Xiaohua Zhai, Thomas Unterthiner, Mostafa Dehghani, Matthias Minderer, Georg Heigold, Sylvain Gelly, et~al.
\newblock An image is worth 16x16 words: Transformers for image recognition at scale.
\newblock \emph{arXiv preprint arXiv:2010.11929}, 2020.

\bibitem[Harisena et~al.(2025)Harisena, Gr{\^e}t-Regamey, and van Strien]{harisena2025}
Nivedita~Varma Harisena, Adrienne Gr{\^e}t-Regamey, and Maarten~J van Strien.
\newblock A novel method to assess spatio-temporal habitat availability for a generalist indicator species group in human-modified landscapes.
\newblock \emph{Landscape Ecology}, 40\penalty0 (6):\penalty0 103, 2025.

\bibitem[He et~al.(2017)He, Gkioxari, Doll{\'a}r, and Girshick]{he2017mask}
Kaiming He, Georgia Gkioxari, Piotr Doll{\'a}r, and Ross Girshick.
\newblock Mask r-cnn.
\newblock In \emph{Proceedings of the IEEE international conference on computer vision}, pp.\  2961--2969, 2017.

\bibitem[Heitzler \& Hurni(2019)Heitzler and Hurni]{heitzler2019}
Magnus Heitzler and Lorenz Hurni.
\newblock Unlocking the geospatial past with deep learning--establishing a hub for historical map data in switzerland.
\newblock \emph{Abstracts of the ICA}, 1:\penalty0 1--2, 2019.

\bibitem[Heitzler \& Hurni(2020)Heitzler and Hurni]{heitzler2020}
Magnus Heitzler and Lorenz Hurni.
\newblock Cartographic reconstruction of building footprints from historical maps: A study on the swiss siegfried map.
\newblock \emph{Transactions in GIS}, 24\penalty0 (2):\penalty0 442--461, 2020.

\bibitem[Houlsby et~al.(2019)Houlsby, Giurgiu, Jastrzebski, Morrone, De~Laroussilhe, Gesmundo, Attariyan, and Gelly]{houlsby2019parameter}
Neil Houlsby, Andrei Giurgiu, Stanislaw Jastrzebski, Bruna Morrone, Quentin De~Laroussilhe, Andrea Gesmundo, Mona Attariyan, and Sylvain Gelly.
\newblock Parameter-efficient transfer learning for nlp.
\newblock In \emph{International conference on machine learning}, pp.\  2790--2799. PMLR, 2019.

\bibitem[Hu et~al.(2022)Hu, Shen, Wallis, Allen-Zhu, Li, Wang, Wang, Chen, et~al.]{hu2022lora}
Edward~J Hu, Yelong Shen, Phillip Wallis, Zeyuan Allen-Zhu, Yuanzhi Li, Shean Wang, Lu~Wang, Weizhu Chen, et~al.
\newblock Lora: Low-rank adaptation of large language models.
\newblock \emph{ICLR}, 1\penalty0 (2):\penalty0 3, 2022.

\bibitem[Jiao et~al.(2022)Jiao, Heitzler, and Hurni]{jiao2022}
Chenjing Jiao, Magnus Heitzler, and Lorenz Hurni.
\newblock A fast and effective deep learning approach for road extraction from historical maps by automatically generating training data with symbol reconstruction.
\newblock \emph{International Journal of Applied Earth Observation and Geoinformation}, 113:\penalty0 102980, 2022.

\bibitem[Ke et~al.(2023)Ke, Ye, Danelljan, Tai, Tang, Yu, et~al.]{ke2023segment}
Lei Ke, Mingqiao Ye, Martin Danelljan, Yu-Wing Tai, Chi-Keung Tang, Fisher Yu, et~al.
\newblock Segment anything in high quality.
\newblock \emph{Advances in Neural Information Processing Systems}, 36:\penalty0 29914--29934, 2023.

\bibitem[Khanam \& Hussain(2024)Khanam and Hussain]{khanam2024yolov11}
Rahima Khanam and Muhammad Hussain.
\newblock Yolov11: An overview of the key architectural enhancements.
\newblock \emph{arXiv preprint arXiv:2410.17725}, 2024.

\bibitem[Kirillov et~al.(2023)Kirillov, Mintun, Ravi, Mao, Rolland, Gustafson, Xiao, Whitehead, Berg, Lo, et~al.]{kirillov2023}
Alexander Kirillov, Eric Mintun, Nikhila Ravi, Hanzi Mao, Chloe Rolland, Laura Gustafson, Tete Xiao, Spencer Whitehead, Alexander~C Berg, Wan-Yen Lo, et~al.
\newblock Segment anything.
\newblock In \emph{Proceedings of the IEEE/CVF international conference on computer vision}, pp.\  4015--4026, 2023.

\bibitem[Li et~al.(2025)Li, Deng, Chen, Meng, Xi, Ma, Wang, Wang, and Zhao]{li2025polyfootnet}
Kai Li, Yupeng Deng, Jingbo Chen, Yu~Meng, Zhihao Xi, Junxian Ma, Chenhao Wang, Maolin Wang, and Xiangyu Zhao.
\newblock Polyfootnet: Extracting polygonal building footprints in off-nadir remote sensing images.
\newblock \emph{IEEE Transactions on Geoscience and Remote Sensing}, 2025.

\bibitem[Lin \& Chiang(2024)Lin and Chiang]{lin2024hyper}
Yijun Lin and Yao-Yi Chiang.
\newblock Hyper-local deformable transformers for text spotting on historical maps.
\newblock In \emph{Proceedings of the 30th ACM SIGKDD Conference on Knowledge Discovery and Data Mining}, pp.\  5387--5397, 2024.

\bibitem[Liu et~al.(2024)Liu, Wang, Yin, Molchanov, Wang, Cheng, and Chen]{liu2024dora}
Shih-Yang Liu, Chien-Yi Wang, Hongxu Yin, Pavlo Molchanov, Yu-Chiang~Frank Wang, Kwang-Ting Cheng, and Min-Hung Chen.
\newblock Dora: Weight-decomposed low-rank adaptation.
\newblock In \emph{Forty-first International Conference on Machine Learning}, 2024.

\bibitem[Mao et~al.(2025)Mao, Xing, Meng, Liu, Bai, Nie, and Meng]{mao2025one}
Xinyu Mao, Xiaohan Xing, Fei Meng, Jianbang Liu, Fan Bai, Qiang Nie, and Max Meng.
\newblock One polyp identifies all: One-shot polyp segmentation with sam via cascaded priors and iterative prompt evolution.
\newblock \emph{arXiv preprint arXiv:2507.16337}, 2025.

\bibitem[Marti-Escofet et~al.(2025)Marti-Escofet, Blumenstiel, Scheibenreif, Fraccaro, and Schindler]{marti2025}
Francesc Marti-Escofet, Benedikt Blumenstiel, Linus Scheibenreif, Paolo Fraccaro, and Konrad Schindler.
\newblock Fine-tune smarter, not harder: Parameter-efficient fine-tuning for geospatial foundation models.
\newblock \emph{arXiv preprint arXiv:2504.17397}, 2025.

\bibitem[Na et~al.(2024)Na, Guo, Jiang, Ma, and Huang]{na2024segment}
Saiyang Na, Yuzhi Guo, Feng Jiang, Hehuan Ma, and Junzhou Huang.
\newblock Segment any cell: A sam-based auto-prompting fine-tuning framework for nuclei segmentation.
\newblock \emph{arXiv preprint arXiv:2401.13220}, 2024.

\bibitem[Raghu et~al.(2021)Raghu, Unterthiner, Kornblith, Zhang, and Dosovitskiy]{raghu2021}
Maithra Raghu, Thomas Unterthiner, Simon Kornblith, Chiyuan Zhang, and Alexey Dosovitskiy.
\newblock Do vision transformers see like convolutional neural networks?
\newblock \emph{Advances in neural information processing systems}, 34:\penalty0 12116--12128, 2021.

\bibitem[R{\"a}th et~al.(2025)R{\"a}th, Gr{\^e}t-Regamey, Xia, Hurni, McPhearson, and van Strien]{rath2025}
Yves~M R{\"a}th, Adrienne Gr{\^e}t-Regamey, Xue Xia, Lorenz Hurni, Timon McPhearson, and Maarten~J van Strien.
\newblock Archetypes of settlement development on the swiss plateau: Identification, description and prediction.
\newblock \emph{Cities}, 159:\penalty0 105791, 2025.

\bibitem[Ravi et~al.(2024)Ravi, Gabeur, Hu, Hu, Ryali, Ma, Khedr, R{\"a}dle, Rolland, Gustafson, et~al.]{ravi2024sam2}
Nikhila Ravi, Valentin Gabeur, Yuan-Ting Hu, Ronghang Hu, Chaitanya Ryali, Tengyu Ma, Haitham Khedr, Roman R{\"a}dle, Chloe Rolland, Laura Gustafson, et~al.
\newblock Sam 2: Segment anything in images and videos.
\newblock \emph{arXiv preprint arXiv:2408.00714}, 2024.

\bibitem[Redmon et~al.(2016)Redmon, Divvala, Girshick, and Farhadi]{redmon2016you}
Joseph Redmon, Santosh Divvala, Ross Girshick, and Ali Farhadi.
\newblock You only look once: Unified, real-time object detection.
\newblock In \emph{Proceedings of the IEEE conference on computer vision and pattern recognition}, pp.\  779--788, 2016.

\bibitem[Ronneberger et~al.(2015)Ronneberger, Fischer, and Brox]{ronneberger2015u}
Olaf Ronneberger, Philipp Fischer, and Thomas Brox.
\newblock U-net: Convolutional networks for biomedical image segmentation.
\newblock In \emph{International Conference on Medical image computing and computer-assisted intervention}, pp.\  234--241. Springer, 2015.

\bibitem[Ryali et~al.(2023)Ryali, Hu, Bolya, Wei, Fan, Huang, Aggarwal, Chowdhury, Poursaeed, Hoffman, et~al.]{ryali2023}
Chaitanya Ryali, Yuan-Ting Hu, Daniel Bolya, Chen Wei, Haoqi Fan, Po-Yao Huang, Vaibhav Aggarwal, Arkabandhu Chowdhury, Omid Poursaeed, Judy Hoffman, et~al.
\newblock Hiera: A hierarchical vision transformer without the bells-and-whistles.
\newblock In \emph{International conference on machine learning}, pp.\  29441--29454. PMLR, 2023.

\bibitem[Shbita et~al.(2020)Shbita, Knoblock, Duan, Chiang, Uhl, and Leyk]{shbita2020building}
Basel Shbita, Craig~A Knoblock, Weiwei Duan, Yao-Yi Chiang, Johannes~H Uhl, and Stefan Leyk.
\newblock Building linked spatio-temporal data from vectorized historical maps.
\newblock In \emph{European semantic web conference}, pp.\  409--426. Springer, 2020.

\bibitem[Shen et~al.(2025)Shen, Zhuang, Kou, Zeng, Xu, and Li]{shen2025mgd}
Haoran Shen, Peixian Zhuang, Jiahao Kou, Yuxin Zeng, Haoying Xu, and Jiangyun Li.
\newblock Mgd-sam2: Multi-view guided detail-enhanced segment anything model 2 for high-resolution class-agnostic segmentation.
\newblock \emph{arXiv preprint arXiv:2503.23786}, 2025.

\bibitem[Sterzinger et~al.(2025)Sterzinger, Peer, and Sablatnig]{sterzinger2025few}
Rafael Sterzinger, Marco Peer, and Robert Sablatnig.
\newblock Few-shot segmentation of historical maps via linear probing of vision foundation models.
\newblock \emph{arXiv preprint arXiv:2506.21826}, 2025.

\bibitem[Sun et~al.(2021)Sun, Hu, Song, and Zhu]{sun2021}
Kai Sun, Yingjie Hu, Jia Song, and Yunqiang Zhu.
\newblock Aligning geographic entities from historical maps for building knowledge graphs.
\newblock \emph{International Journal of Geographical Information Science}, 35\penalty0 (10):\penalty0 2078--2107, 2021.

\bibitem[Wang et~al.(2022)Wang, Zheng, Chen, and Wang]{wang2022fourier}
Peihao Wang, Wenqing Zheng, Tianlong Chen, and Zhangyang Wang.
\newblock Anti-oversmoothing in deep vision transformers via the fourier domain analysis: From theory to practice.
\newblock \emph{arXiv preprint arXiv:2203.05962}, 2022.

\bibitem[Wang et~al.(2024)Wang, Misra, Zeng, Girdhar, and Darrell]{wang2024}
Xudong Wang, Ishan Misra, Ziyun Zeng, Rohit Girdhar, and Trevor Darrell.
\newblock Videocutler: Surprisingly simple unsupervised video instance segmentation.
\newblock In \emph{Proceedings of the IEEE/CVF Conference on Computer Vision and Pattern Recognition}, pp.\  22755--22764, 2024.

\bibitem[Wu et~al.(2023)Wu, Zhang, and Elbatel]{wu2023self}
Qi~Wu, Yuyao Zhang, and Marawan Elbatel.
\newblock Self-prompting large vision models for few-shot medical image segmentation.
\newblock In \emph{MICCAI workshop on domain adaptation and representation transfer}, pp.\  156--167. Springer, 2023.

\bibitem[Xia et~al.(2022)Xia, Heitzler, and Hurni]{xia2022cnn}
Xue Xia, Magnus Heitzler, and Lorenz Hurni.
\newblock Cnn-based template matching for detecting features from historical maps.
\newblock \emph{International Archives of the Photogrammetry, Remote Sensing and Spatial Information Sciences}, 43:\penalty0 1167--1173, 2022.

\bibitem[Xia et~al.(2023)Xia, Jiao, and Hurni]{xia2023contrastive}
Xue Xia, Chenjing Jiao, and Lorenz Hurni.
\newblock Contrastive pretraining for railway detection: Unveiling historical maps with transformers.
\newblock In \emph{Proceedings of the 6th ACM SIGSPATIAL International Workshop on AI for Geographic Knowledge Discovery}, pp.\  30--33, 2023.

\bibitem[Xia et~al.(2024{\natexlab{a}})Xia, Zhang, Heitzler, and Hurni]{xia2024vectorizing}
Xue Xia, Tao Zhang, Magnus Heitzler, and Lorenz Hurni.
\newblock Vectorizing historical maps with topological consistency: A hybrid approach using transformers and contour-based instance segmentation.
\newblock \emph{International Journal of Applied Earth Observation and Geoinformation}, 129:\penalty0 103837, 2024{\natexlab{a}}.

\bibitem[Xia et~al.(2024{\natexlab{b}})Xia, Zhang, and Hurni]{xia2024video}
Xue Xia, Tao Zhang, and Lorenz Hurni.
\newblock Video instance segmentation is all you need for linking geographic entities from historical maps.
\newblock In \emph{IGARSS 2024-2024 IEEE International Geoscience and Remote Sensing Symposium}, pp.\  8491--8494. IEEE, 2024{\natexlab{b}}.

\bibitem[Xia et~al.(2025)Xia, Zhang, Song, Huang, and Hurni]{xia2025mapsam}
Xue Xia, Daiwei Zhang, Wenxuan Song, Wei Huang, and Lorenz Hurni.
\newblock Mapsam: adapting segment anything model for automated feature detection in historical maps.
\newblock \emph{GIScience \& Remote Sensing}, 62\penalty0 (1):\penalty0 2494883, 2025.

\bibitem[Yan et~al.(2023)Yan, Li, Li, Zhou, Zhang, Feng, Diao, Fu, and Sun]{yan2023ringmo}
Zhiyuan Yan, Junxi Li, Xuexue Li, Ruixue Zhou, Wenkai Zhang, Yingchao Feng, Wenhui Diao, Kun Fu, and Xian Sun.
\newblock Ringmo-sam: A foundation model for segment anything in multimodal remote-sensing images.
\newblock \emph{IEEE Transactions on Geoscience and Remote Sensing}, 61:\penalty0 1--16, 2023.

\bibitem[Zhang \& Liu(2023)Zhang and Liu]{zhang2023customized}
Kaidong Zhang and Dong Liu.
\newblock Customized segment anything model for medical image segmentation.
\newblock \emph{arXiv preprint arXiv:2304.13785}, 2023.

\bibitem[Zhang et~al.(2023)Zhang, Jiang, Guo, Yan, Pan, Ma, Dong, Gao, and Li]{zhang2023personalize}
Renrui Zhang, Zhengkai Jiang, Ziyu Guo, Shilin Yan, Junting Pan, Xianzheng Ma, Hao Dong, Peng Gao, and Hongsheng Li.
\newblock Personalize segment anything model with one shot.
\newblock \emph{arXiv preprint arXiv:2305.03048}, 2023.

\bibitem[Zhu et~al.(2024)Zhu, Hamdi, Qi, Jin, and Wu]{zhu2024}
Jiayuan Zhu, Abdullah Hamdi, Yunli Qi, Yueming Jin, and Junde Wu.
\newblock Medical sam 2: Segment medical images as video via segment anything model 2.
\newblock \emph{arXiv preprint arXiv:2408.00874}, 2024.

\end{thebibliography}
\bibliographystyle{iclr2025_conference}

% \appendix
% \section{Appendix}

\end{document}